\documentclass[a4paper,11pt]{article}
\usepackage[ruled]{algorithm2e}
\usepackage{algorithmic}
\usepackage{amssymb,,amsmath,epsfig,amsthm,authblk,dsfont,bm}
\usepackage{enumitem}
\usepackage{smile}
\usepackage[numbers,sort&compress]{natbib}
\usepackage[margin=1in]{geometry}
\usepackage{setspace,color}
\usepackage{tikz}
\usepackage{epstopdf}
\usepackage{grffile}
\usepackage{makecell}
\usepackage{multirow}
\usepackage{graphbox}
\usepackage{lineno}
\usepackage{hyperref}
\hypersetup{
	colorlinks=true,
	linkcolor=blue,
	citecolor=blue,
	filecolor=blue,      
	urlcolor=blue
}

\providecommand{\keywords}[1]{\textbf{Key words:} #1}

\newcommand{\Sig}{\mathrm{Sig}}
\newcommand{\ReLU}{\mathrm{ReLU}}

\allowdisplaybreaks
\begin{document}
	\title{
A Mathematical Explanation of UNet
 } 
\author{
Xue-Cheng Tai\thanks{Norwegian Research Centre (NORCE), Nyg{\aa}rdsgaten 112, 5008 Bergen, Norway.  Email: xtai@norceresearch.no, xuechengtai@gmail.com. The work of Xue-Cheng Tai is partially supported by HKRGC-NSFC Grant N-CityU214/19,  HKRGC CRF Grant C1013-21GF and NORCE Kompetanseoppbygging program.},
Hao Liu\thanks{Department of Mathematics, Hong Kong Baptist University, Kowloon Tong, Hong Kong SAR. Email: haoliu@hkbu.edu.hk. The work of Hao Liu is partially supported by National Natural Science Foundation of China 12201530 and HKRGC ECS 22302123.},
Raymond H. Chan\thanks{Lingnan University, Tuen Mun, Hong Kong SAR. Email: raymond.chan@ln.edu.hk. The work of Raymond H. Chan is partially supported by
HKRGC GRF grants CityU1101120, CityU11309922, CRF grant C1013-21GF, and HKRGC-NSFC Grant N-CityU214/19.},
Lingfeng Li\thanks{Hong Kong Center for Cerebro-Cardiovascular Health Engineering, Hong Kong SAR. The work of Lingfeng Li is supported by the InnoHK project at Hong Kong Centre for Cerebro-Cardiovascular Health Engineering (COCHE). Email: lfli@hkcoche.org }

 }
	\date{ }

\maketitle

\begin{abstract}
The UNet architecture has transformed image segmentation.
UNet's versatility and accuracy have driven its widespread adoption, significantly advancing fields reliant on machine learning problems with images.
In this work, we give a clear and concise mathematical explanation of UNet. 
We explain what is the meaning and function of each of the components of UNet. 
We will show that UNet is solving a control problem. We decompose the control variables using multigrid methods. Then, operator-splitting techniques is used to solve the problem, whose architecture exactly recovers the UNet architecture. Our result shows that UNet is a one-step operator-splitting algorithm for the control problem.
\end{abstract}


\section{Introduction}

Deep neural networks have made remarkable successes in many tasks, including image segmentation \cite{ronneberger2015u,long2015fully,  liu2022deep,zhou2019unetpp}, image denoising \cite{zhang2017beyond,anwar2019real,wu2024medical}, image classification, natural language processing \cite{graves2013speech}, etc. Among these works, UNet \cite{ronneberger2015u} stands out as a renowned network and inspired a lot of following works \cite{chen2017deeplab, badrinarayanan2017segnet,zhou2019unetpp,xu2024mef}.

UNet was originally proposed for medical image segmentation. It consists of four components: encoder, decoder, bottleneck and skip-connections. Given an input image, the encoder part conducts dimension reduction and convert the image to a low-dimensional tensor. The bottleneck performs some operations on the tensor, after which the tensor is converted to the segmented image by the decoder. Skip-connections are used to directly pass information from encoder to decoder. UNet does a great job in medical image segmentation, and has garnered significant attention. Its encoder-decoder architecture inspired a lot of subsequent works, including DeepLab \cite{chen2017deeplab}, SegNet \cite{badrinarayanan2017segnet}, UNet++ \cite{zhou2019unetpp} for image segmentation, SUNet \cite{fan2022sunet}, RDUNet \cite{gurrola2021residual} for image denoising.

A series of works have been aimed at elucidating the empirical successes of deep neural networks \cite{barron1993universal,yarotsky2017error,zhou2020universality,chen2019efficient} and establishing connections between deep learning and mathematical models \cite{Weinan2017, ruthotto2020deep,tan2022unsupervised,liu2022deep}. 
The current work is inspired by a series of earlier researches. In \cite{Weinan2017, Weinan2020},  the authors initiated the idea to treat networks as discretized representations of continuous dynamical systems.
 The authors of \cite{benning2019deep}  studied the connections between networks and control problems. PDE and ODE-motivated stable network architectures are proposed in \cite{ruthotto2020deep,Haber2018}. Inspired by the weak formulation of PDEs, \cite{zang2020weak} proposed weak adversarial networks for solving PDEs. This idea was further applied in \cite{bao2024wanco} to solve constrained optimization problems. Many networks are designed with an encoder-decoder architecture, in which the encoder and decoder are expected to extract and reconstruct features of data, respectively. Analogies between this architecture and multiscale methods are pointed out in \cite{Haber2018b, Haber2018}. In \cite{He2019a}, the authors proposed to use operators with multigrid methods to extract and reconstruct features. In \cite{lan2022dosnet}, the authors used networks based on the operator-splitting method to solve PDEs. For image processing, the regularizers of prior information are incorporated with networks to design new networks. Networks with volume-preserving properties and star-shape priors are proposed in \cite{liu2022deep} for image segmentation. Compactness priors are used in \cite{zhang2024deep}. In \cite{tan2022unsupervised}, a  multi-task deep variational model is proposed which variational models are incorporated into the loss functions. Based on the Chan-Vese model \cite{chan2001active} and fields of experts regularizer, a novel deep neural network is proposed in \cite{cui2024trainable} for image segmentation.

Based on the Potts model and operator-splitting methods, networks with mathematical explanations are proposed in \cite{tai2024pottsmgnet,liu2023connections,liu2023double}.
In  \cite{tai2024pottsmgnet}, the authors proposed PottsMGNet by integrating the Potts model, operator-splitting method, control problem, and multigrid method, which provides a mathematical explanation of the encoder-decoder-based networks. PottsMGNet demonstrated great performances in segmenting images with various noise levels using a single network. It was shown in \cite{tai2024pottsmgnet} that most of the encoder-decoder-based neural networks are essentially operator-splitting algorithms solving certain control problems. The double-well net proposed in \cite{liu2023double} utilizes the Potts model, operator-splitting methods, the double-well potential, and network representation theories. In double-well nets, a network is used to represent the region force term in the Potts model, providing a data-driven way to learn the region force term.
The works mentioned above make connections among mathematical models, algorithms, and deep neural networks. However, the resulting networks are more or less different from UNet and cannot be directly applied to provide an explanation of UNet. 
In this paper, we aim to provide a clear and concise mathematical explanation of UNet.
Building on the key concepts from \cite{tai2024pottsmgnet}, we rigorously formulate the problem to show that the network derived from the splitting-multigrid algorithms for the control problem corresponds exactly to UNet when only a single iteration of the algorithm is applied. In fact, UNet emerges as a special case of the more general algorithm described in \cite{tai2024pottsmgnet}.
The central ideas for multigrid methods we use in this work for solving minimization problems come from \cite{tai2003rate,tai1998subspace,tai2002global,xu1992iterative}. The general explanations and convergence proofs provided in these works for multigrid methods present the method in a more general form, encompassing linear elliptic solvers as special cases and suits our proposed control problem well. 
Operator-splitting methods decompose complicated problems into multiple easy-to-solve sub-problems and are widely used in solving PDEs \cite{glowinski2019finite}, inverse problems \cite{glowinski2015penalization} and image processing \cite{deng2019new,liu2021color,duan2022fast}. We suggest readers to \cite{glowinski2016some,glowinski2017splitting,glowinski2019fast} for a comprehensive discussion on operator-splitting methods.  Traditional splitting methods decompose the original problem into a small number of sub-problems. In  the context of this work, the number of the decomposed sub-problems are rather large and thus need to introduce some hybrid splitting schemes as in Section \ref{sec.hybrid} proposed in \cite{tai2024pottsmgnet}

In this work, starting from a control problem, we will first derive its equivalent problem by introducing an indicator function. We then use the multigrid idea to decompose the control variables into different scales and utilize the hybrid splitting strategy to propose an operator-splitting method for the new problem. The algorithm consists of several sub-steps, each of which contains an explicit linear convolution step and an implicit step, where the implicit step has a closed-form solution which turns out to be the ReLU function. We show that the resulting algorithm exactly recovers the UNet architecture. Our results show that UNet is a one-step operator-splitting algorithm solving a control problem.
 
This paper is organized as follows: In Section \ref{sec.formulation}, we present the control problem, derive its equivalent form using an indicator function, and introduce basic ideas of hybrid operator-splitting methods and multigrid methods. We discuss in Section \ref{sec.alg} the decomposition of control variables and present our proposed operator-splitting method to solve the control problem. Solutions to subproblems in the proposed algorithm are presented in Section \ref{sec.implementation}. We discuss connections between the proposed algorithm and general networks and how the proposed algorithm recovers UNet in Section \ref{sec.networks}, and conclude this paper in Section \ref{sec.conclusion}.

\section{Proposed formulation} \label{sec.formulation}
In this section, we present our control problem and briefly introduce hybrid operator-splitting methods and multigrid methods.
 \subsection{The control problem}
 Given an input image $f$, we consider the following initial value problem
 \begin{equation}
	\begin{cases}
		\frac {\partial u(\xb,t)}{\partial t}  =  W(\xb,t) \ast u(\xb,t) +  d(t) -\ln  \frac {u(\xb,t)} {1-u(\xb,t)} , 
		\ (\xb,t) \in \Omega\times (0,T], \\ 
		u(\xb, 0)      = H(f), \ \xb \in \Omega,
	\end{cases}
	\label{eq.control0}
\end{equation}
where $W(\xb,t),  d(t)$ are control variables that governs the dynamics of $u$, $\ast$ denotes convolution, $H(f)$ is some operation to generate initial condition from $f$, $\Omega$  is the domain where the image is defined and $T$ is some fixed time. Due to the appearance of the term $\ln  \frac u {1-u}$, the solution of the above equation is forced to be  in $(0,1)$. For numerical consideration and to make the connection between operator-splitting methods and neural networks clearer, we introduce a constraint and consider the following constrained control problem
 \begin{equation}
	\begin{cases}
		\frac {\partial u}{\partial t}  =  W(\xb,t) \ast u(\xb,t) +  d(t) -\ln  \frac {u(\xb,t)} {1-u(\xb,t)} , 
		\ (\xb,t) \in \Omega\times (0,T], \\ 
		u(\xb,t) \geq0,\\
		u(\xb, 0)      = H(f), \ \xb \in \Omega. 
	\end{cases}
	\label{eq.control.constraint}
\end{equation}
Due to the property of the term $\ln  \frac u {1-u}$, the introduced constraint does not change the solution.

Next, we incorporate the constraint into the equation by introducing an indicator function. This technique has been used in designing fast operator-splitting methods for image processing \cite{deng2019new,liu2021color,liu2022operator,duan2022fast}. Define the set 
$$
\Sigma=\{u: u(\xb,t)\geq 0 \mbox{ for } (\xb,t)\in \Omega\times (0,T]\}
$$
and its indicator function
$$
\cI_{\Sigma}(u)=\begin{cases}
	0 & \mbox{ if } u\in \Sigma,\\
	\infty & \mbox{ otherwise}.
\end{cases}
$$
Problem (\ref{eq.control.constraint}) is equivalent to the following unconstrained control problem
 \begin{equation}
	\begin{cases}
		\frac {\partial u}{\partial t} - W(\xb,t) \ast u - d(t) +\ln  \frac u {1-u} +\partial \cI_{\Sigma}(u) \ni 0,
		\ (\xb,t) \in \Omega\times (0,T], \\ 
		u(\xb, 0)      = H(f), \ \xb \in \Omega,
	\end{cases}
	\label{eq.control}
\end{equation}
where $\partial\cI_{\Sigma}$ denotes the subdifferential of $\cI_{\Sigma}$.

By solving (\ref{eq.control}) for any input image $f$, We expect that $u(\xb,0)$ will evolve to $u(\xb,T)$ which is close to a binary function. 
For a given dataset $\{f_i,g_i\}_{i=1}^I$,  we consider a control problem. Specifically, denote $\theta_1=\{W(\xb,t),d(t)\}$ as the set of control variables, and $\cN_1:f\rightarrow u(\xb,T)$ as the mapping from $f$ to the solution of  (\ref{eq.control}) at time $T$: $\cN_1(f;\theta_1)=u(\xb,T)$. We optimize $\theta_1$ by solving
    \begin{equation}\label{eq:loss}
		\min_{\theta_1} \sum_{i=1}^I \mathcal{L}(\mathcal{N}_1(f_i,\theta_1) , g_i ),
	\end{equation}
 where $\cL(\cdot,\cdot)$ is the loss function measuring the differences between its arguments. Common loss functions include logistic loss and hinge loss.

 \subsection{Hybrid splitting methods}
 \label{sec.hybrid}
 We will use the hybrid splitting method proposed in \cite{tai2024pottsmgnet} to solve (\ref{eq.control}).  Refer to \cite{glowinski2016some,glowinski2017splitting,glowinski2019fast} for some general introduction to traditional splitting methods. In this subsection, we give a brief introduction to the hybrid splitting method. Consider a general initial value problem
	\begin{align}
		\begin{cases}
			u_t+\displaystyle\sum_{m=1}^M \left(\sum_{k=1}^{c_m} \sum_{s=1}^{c_{m-1}} A_{k,s}^m(\xb,t;u) +\sum_{k=1}^{c_m} S_{k}^m(\xb,t;u) +\sum_{k=1}^{c_m}	f_{k}^m(\xb,t)\right)=0 \mbox{ on } \Omega\times [0,T],\\
			u(\xb,0)=u_0(\xb),
		\end{cases}
		\label{eq.general.hybrid}
	\end{align}
	where $\{c_m\}_{m=0}^M$ are some given positive integers with $c_0=c_M=1$, $A_{k,s}^m, S_k^m$ are operators, $f_k^m$'s are some given  functions independent of $u$.   The hybrid splitting method is a mixture of sequential splitting and parallel splitting. Briefly speaking, the hybrid splitting method arranges parallel splittings sequentially.

\begin{figure}[t!]
		\centering
		\includegraphics[width=0.9\textwidth]{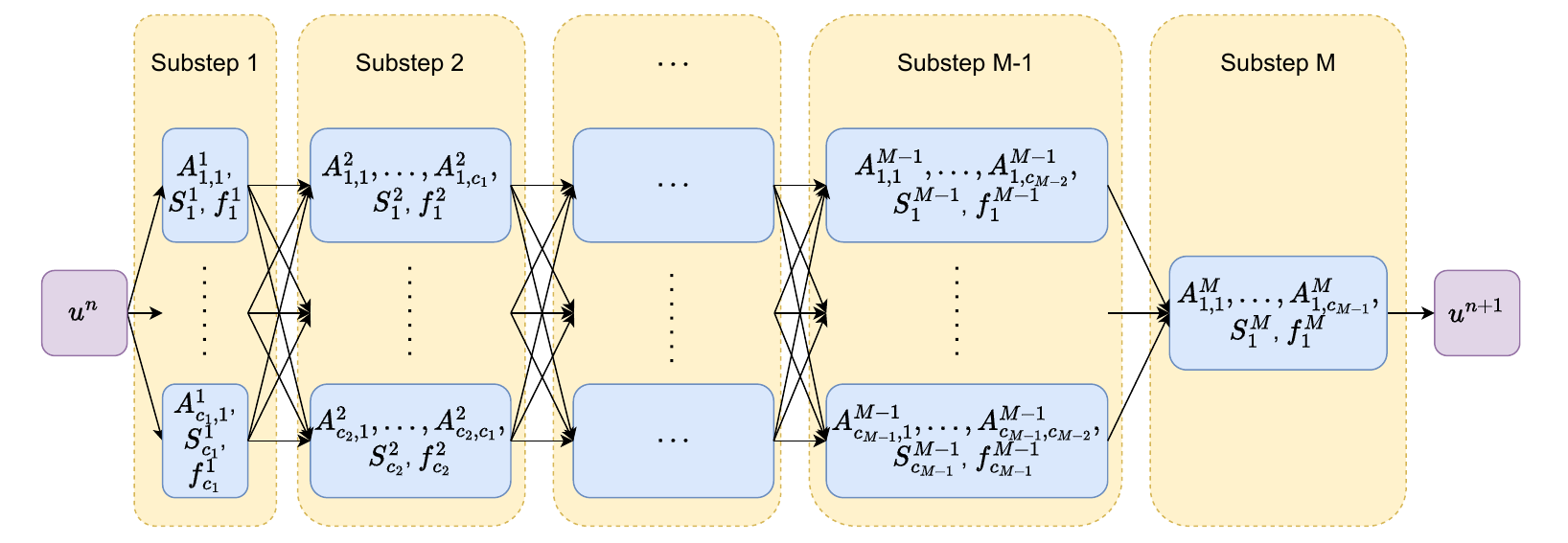}
		\caption{An illustration of Algorithm \ref{alg.hybrid}.}
		\label{fig.hybrid}
	\end{figure}

    The algorithm of hybrid splitting is summarized in Algorithm \ref{alg.hybrid}. In the algorithm, all operators are distributed into $M$ sequential sub-steps, each of which is a parallel splitting with $c_{m}$ parallel pathways. The computation of each parallel pathway uses the $c_{m-1}$ intermediate results from the previous sub-step. The structure of Algorithm \ref{alg.hybrid} is illustrated in Figure \ref{fig.hybrid}.
	
	\begin{algorithm}[th!]
		\caption{A hybrid splitting scheme}\label{alg.hybrid}
		\begin{algorithmic}
			\STATE {\bf Data:} The solution $u^n$ at time step $t^n$.
			\STATE {\bf Result:} The computed solution $u^{n+1}$ at time step $t^{n+1}$.
			
			{\bf Set} $d_{1} = 1,u_1^{n} = u^n$. \\
			\FOR {$m=1,...,M$}
			\FOR {$k=1,...,c_{M}$}
			\STATE Compute $u^{n+m/M}_k$ by solving
			\begin{align}
				\frac {u^{n+m/M}_k - u^{n+(m-1)/M}} {c_m \Delta t } =   
				-\sum_{s=1}^{c_{m-1}} 
				&A_{k,s}^{m}( t^n;u_s^{n+(m-1)/M}) \nonumber\\
                &- S_k^m(t^{n+1};u^{n+m/M})-f_k^m(t^n).
				\label{eq.hybrid.k}
			\end{align}
			
			\ENDFOR
			
			Compute $u^{n+m/M}$ as
			\begin{equation}
				u^{n+m/M} = \frac 1 {c_m} \sum_{k=1}^{c_m} u_k^{n+m/M}. 
				\label{eq.hybrid.ave}
			\end{equation}
			\ENDFOR
		\end{algorithmic}
	\end{algorithm}
 
    The above scheme splits the original problem into $M$ sequential steps with $m=1,2,\cdots. M$. Inside each sequential step,  the problem is further split into  $c_m$ parallel steps for each $m$.  For each of these parallel subproblems, we treat the operator $S_k^m$ using implicit approximation and the operators $A_{k,s}^m$ using explicit approximations.   
    It is shown in \cite{tai2024pottsmgnet} that when all operators in (\ref{eq.general.hybrid}) are linear, Algorithm \ref{alg.hybrid} converges with first-order accuracy: 
	\begin{theorem}[Theorem D.1 in \cite{tai2024pottsmgnet}]\label{thm.hybrid}
		For a fixed $T>0$ and a positive integer $N$, set $\Delta t=T/N$. Let $u^{n+1}$ be the numerical solution by Algorithm \ref{alg.hybrid}. Assume $A_{k,s}^m$'s and $S_k^m$'s are Lipschitz with respect to $t,\xb$, and are linear symmetric positive definite operators with respect to $u$. Assume $\Delta t$ is small enough (i.e., $N$ is large enough). We have 
		\begin{align}
			\|u^{n+1}-u(t^{n+1})\|_{\infty}=O(\Delta t)
		\end{align}
		for any $0\leq n\leq N$.
	\end{theorem}
	   In applying this algorithm to our control problem, the $A_{k,s}^m$ operators  are coming from the decomposed control variables  which are the convolutional kernels over the different levels of the multigrids explained in the next sections.  
	
	\subsection{Multigrid discretizations}
 \label{sec.multigrid}
 To demonstrate our splitting strategy, we will use the multigrid idea to decompose the control variables into components with different scales. 
 in this subsection, we present the multigrid method for a general function $f$, which we will refer to as an image to remain consistent with the terminology.
 
  Denote the original resolution (finest grid) of an image $f$ by $\cT$ with size $m\times n$, and grid step size $h$, with 
	$$m=2^{s_1},\quad n=2^{s_2}$$
	for some $h>0$ and integers $s_1,s_2>0$. The image $f$ is considered to have a constant value on each element (or called pixel) $[\alpha_1h,(\alpha_1+1)h)\times[\alpha_2h,(\alpha_2+1)h)$ for $\alpha_1=0,...,m-1$ and $\alpha_2=0,...,n-1$. 

    Set $\cT^1=\cT$. Given grid $\cT^j$, for the next level coarse grid $\cT^{j+1}$, we downsample the number of grid points along each dimension by half. Following this process, we can generate a sequence of grids $\{\cT^j\}_{j=1}^J$ with $J$ denoting the coarsest level of grids and each $\cT^j$ has grid size $m_j\times n_j$ and grid step size $h_j$ with
	\begin{align*}
		&	m_j=2^{s_1-j+1},\quad n_j=2^{s_2-j+1}, \quad h_j=2^{j-1}h.
	\end{align*}
	
	Denote 
	$
	\cI^j=\{ \balpha  : \balpha= (\alpha_1,\alpha_2), \alpha_1=0,...,m_j-1,\ \alpha_2=0,...,n_j-1\}. 
	$
	For a given grid $\cT^j$, a set of piecewise-constant basis functions
	$\{\phi^j_{\alpha} \}_{\alpha \in I^j}$ is defined as
	\begin{align}
		\phi^j_{\balpha}(x,y)=\begin{cases}
			1 & \mbox{ if } (x,y)\in [\alpha_1h_j,(\alpha_1+1)h_j)\times[\alpha_2h_j,(\alpha_2+1)h_j),\\
			0 & \mbox{ otherwise}.
		\end{cases}
		\label{eq.basis}
	\end{align}
	Let $\cV^j ={\rm span}(\{\phi^j_{\alpha} \}_{\alpha \in \cI^j})$ be the linear space containing all the piecewise constant functions over grid $\cT^j$, we have
	\begin{align}
		\cV^1\supset \cV^2 \supset \cdots \supset \cV^J. 
		\label{eq.fspace.relation}
	\end{align}
	For each $f\in \cV^j$, it can be expressed as
	$f(x,y)=\sum_{\balpha\in \cI^j} f_{\balpha}^j\phi_{\balpha}^j(x,y)$
	with $f_{\balpha}^j=f(\alpha_1h_j,\alpha_2h_j)$.
	
	Next, we introduce the downsampling and upsampling operations that convert functions between different grids.  Let $\cT^j$ and $\cT^{j+1}$ be two grids. Consider $f^{j+1}\in \cV^{j+1}$.  According to (\ref{eq.fspace.relation}), there exists a function $f^{j}\in \cV^{j}$ satisfying $f^{j}=f^{j+1}$. Denote the upsampling operator $\cU:\cV^{j+1} \rightarrow \cV^{j} $ for any $j>0$ so that
	\begin{align}
		f^j=\cU(f^{j+1}).
	\end{align}
	One can show that for $\balpha\in \cI^j$, it holds
	\begin{align}
		(\cU(f^j))_{\balpha}=f^{j+1}_{\balpha'} \mbox{ with } \alpha_1',\alpha_2' \mbox{ satisfying } 2\alpha_1'-1\leq \alpha_1 \leq 2\alpha_1', \  2\alpha_2'-1\leq \alpha_2 \leq 2\alpha_2'.
		\label{eq.upsampling}
	\end{align} 
	The mapping discussed above is the simplest upsampling operator. One can also choose other upsampling operators that apply some operations while upsampling, such as interpolation or transpose convolution.

	For the downsampling operator $\cD^j: \cV^{j} \rightarrow \cV^{j+1}$, there are many ways to define it. For example, given a function $f^{j}\in \cV^j$,  we can define $\cD^j$ as an averaging downsampling operator:
	\begin{align}
		f^{j+1}=(\cU^j(f^{j}))_{\balpha}= \frac{1}{4}\sum_{\alpha_1'=2\alpha_1-1}^{2\alpha_1} \sum_{\alpha_2'=2\alpha_2-1}^{2\alpha_2} f^j_{\alpha_1',\alpha_2'}.
		\label{eq.downsampling.ave}
	\end{align} 
	Another choice is the max pooling operator which is widely used in deep learning:
	\begin{align}
		f^{j+1}=(\cU^k(f^{j}))_{\balpha}= \max_{\substack{\alpha_1'=2\alpha_1-1,2\alpha_1\\ \alpha_2'=2\alpha_2-1,2\alpha_2}} f^j_{\alpha_1',\alpha_2'}.
		\label{eq.downsampling.max}
	\end{align}

	\begin{figure}[t!]
		\centering
		\includegraphics[width=0.6\textwidth]{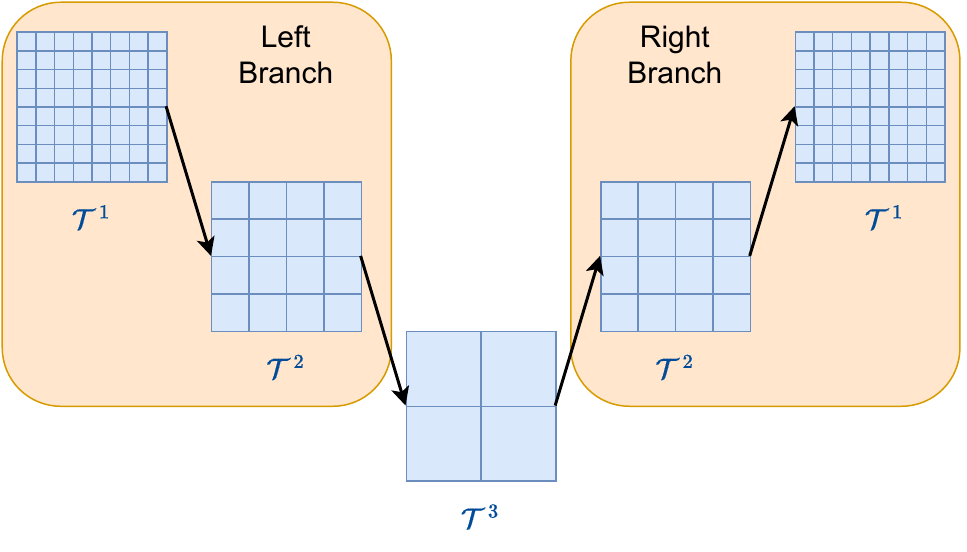}
		\caption{An illustration of a V-cycle of the multigrid method.}
		\label{fig.V}
	\end{figure}

\section{The proposed algorithm}
\label{sec.alg}
We decompose the control variables in (\ref{eq.control}) using the multigrid idea and then propose an algorithm based on the hybrid operator-splitting method to solve it. After these, it will then be shown that UNet is exactly one-step of the operator-splitting algorithm for the control problem.

 \subsection{Decomposition of control variables $\theta_1$}
In traditional multigrid methods, a popular framework is the "fine-grid $\rightarrow$ coarse grid $\rightarrow$ fine grid" strategy \cite{chan1994domain}. Such forms of V-cycle multigrid method can be interpreted as space decomposition and subspace correction \cite{tai2003rate,tai1998subspace,tai2002global,xu1992iterative}, see Figure \ref{fig.V} for an illustration. Traditional multigrid methods solve the decomposed subproblems by simple Gauss-Seidel or Jacobi iterations.  In our approach shown here,  we solve the subproblems by operating splitting  sequentially or in parallel over the decomposed function subspaces. 

We will decompose $\theta_1$ into a sum of variables with different scales over the multigrids. Then, we use a hybrid splitting method to solve (\ref{eq.control}) so that all decomposed variables are distributed into several subproblems, which are solved sequentially or in parallel. Within one iteration of the splitting method, all decomposed variables are gone through.
The general splitting idea is to split the operators based on a V-cycle according to the scale level. We assign several sub-steps to each scale level of each branch of the V-cycle. Each sub-step consists of several parallel splitting pathways.

We decompose all terms in the right-hand side of (\ref{eq.control}) via the following six steps:
\begin{enumerate}[label=(\roman*)]
    \item According to the idea of a V-cycle, we decompose $W(\xb,t)$ and $d(t)$ as
    \begin{align}\label{eq:decomp}
			W(\xb,t)=A(\xb,t)+\tilde{A}(\xb,t), \ d(t)=b(t)+\tilde{b}(t).
		\end{align}
		These variables will be further decomposed next. Above, $A, b$ are sums of control variables in the left branch of the V-cycle, and $\tilde{A}, \tilde b$ are sums of the control variables in the right branch. We also decompose the nonlinear operator as follows: 
		\begin{align}
			- \ln  \frac u {1-u} - \partial \cI(u)=S(u)+\tilde{S}(u).
			\label{eq.S.sum}
		\end{align}
  Here $S(u)$ contains nonlinear operations in the left branch and $\tilde{S}(u)$ contains nonlinear operations in the right branch. In particular, we put $- \ln  \frac u {1-u}$ in $\tilde{S}(u)$ only, i.e., $S(u)$ only contains operator $\partial \cI(u)$. Later, we will show our operator splitting method recovers UNet. The operation $- \ln  \frac u {1-u}$ corresponds to the sigmoid layer at the end of UNet.

  \item We further decompose the operators into components at different scales as: 
    \begin{align}
        &A(\xb,t)=\sum_{j=1}^J A^j(\xb,t),\quad b(t)=\sum_{j=1}^{J-1} b^j(t), \quad S(u)=\sum_{j=1}^J S^j(u),\\
        &\tilde{A}(\xb,t)=\sum_{j=1}^{J-1} \tilde{A}^j(\xb,t)+A^*(\xb,t),	\quad \tilde{b}(t)=\sum_{j=1}^{J-1} \tilde{b}^j(t)+ b^*(t),\quad \tilde{S}(u)=\sum_{j=1}^{J-1} \tilde{S}^j(u)+S^*(u),
        \label{eq.decompose.2}
    \end{align}
    where $A^j,b^j,\widetilde{A}^j, \widetilde{b}^j$ contain control variables at grid level $j$, $A^*$, $b^*$ are control variables that are applied to the output of the V-cycle at the finest mesh, i.e. $A^j, \tilde A^j\in \cV^j, A^*\in \cV^1, b^j, \tilde b^j, b^* \in \mathbb{R}$.
    Operators $S^j,\widetilde{S}^j$ are applied to the intermediate solution on grid level $j$. Operator $S^*$ is applied to the output of the V-cycle at the finest mesh.

   \item At grid level $j$, let $L_j$ be the number of sub-steps to be solved at grid level $j$ in the left and right branches of the V-cycle. We decompose 
    \begin{align}
        A^{j}(\xb,t)=\sum_{l=1}^{L_j} A^{j,l}(\xb,t), \quad b^j(t)=\sum_{l=1}^{L_j} b^{j,l}(t),\quad  S^j(u)=\sum_{l=1}^{L_j} S^{j,l}(u),\\
        \widetilde{A}^{j}(\xb,t)=\sum_{l=1}^{L_j} \widetilde{A}^{j,l}(\xb,t), \quad \widetilde{b}^j(t)=\sum_{l=1}^{L_j} \widetilde{b}^{j,l}(t),\quad  \widetilde{S}^j(u)=\sum_{l=1}^{L_j} \widetilde{S}^{j,l}(u).
    \end{align}
    In our splitting scheme, we will use a sequential splitting techniques for the operators given above both for the left and right branch, where $A^{j,l},b^{j,l}$, and $S^{j,l}$ are the operators at the $l$-th sequential sub-step of the left branch, $\widetilde{A}^{j,l},\widetilde{b}^{j,l}$ and $\widetilde{S}^{j,l}$ are the operators  at the $l$-th sequential sub-step of the right branch.
    
    \item At grid level $j$, for each sequential sub-step $l$ of each branch, we decompose 
    \begin{align}
        &A^{j,l}(\xb,t)=\sum_{k=1}^{c_j} A^{j,l}_k(\xb,t), \quad b^{j,l}(t)=\sum_{k=1}^{c_j} b^{j,l}_k(t), \quad  S^{j,l}(u)=\sum_{k=1}^{c_j} S^{j,l}_k(u),\\
        &\widetilde{A}^{j,l}(\xb,t)=\sum_{k=1}^{c_j} \widetilde{A}^{j,l}_k(\xb,t), \quad \widetilde{b}^{j,l}(t)=\sum_{k=1}^{c_j} \widetilde{b}^{j,l}_k(t), \quad  \widetilde{S}^{j,l}(u)=\sum_{k=1}^{c_j} \widetilde{S}^{j,l}_k(u).
    \end{align}
    At grid level $j$, we split these operators into $c_j$ parallel pathways, where $A^{j,l}_k,b^{j,l}_k$ and $S^{j,l}_k$ are used in the $k$-th parallel splitting pathway in the left branch, $\widetilde{A}^{j,l}_k,\widetilde{b}^{j,l}_k$ and $\widetilde{S}^{j,l}_k$ are used in the $k$-th parallel splitting pathway in the right branch. 
    \item For the left branch, at grid level $j$, the $l$-th sequential step and the $k$-th parallel splitting pathway, we take all $c_{j-1}$ outputs from the previous sequential step as inputs and use components from $A^{j,l}_k$ to convolve with them. We decompose $A^{j,l}_k$ into $c_{j-1}$ kernels:
    \begin{align}
        A^{j,l}_k(\xb,t)=\sum_{s=1}^{c_{j-1}} A^{j,l}_{k,s}(\xb,t) \quad \mbox{ with } \quad 
        c_{j,l}=\begin{cases}
            c_{j-1} & \mbox{ if } l=1,\\
            c_j & \mbox{ if } l>1.
        \end{cases}
        \label{eq.c}
    \end{align} 
    Similarly, for the right branch, the previous sub-step has $c_{j+1}$ outputs. We decompose $\widetilde{A}^{j,l}_k$ into $c_{j+1}$ kernels:
    \begin{align}
        \widetilde{A}^{j,l}_k(\xb,t)=\sum_{s=1}^{\widetilde{c}_j} \widetilde{A}^{j,l}_{k,s}(\xb,t) \quad \mbox{ with } \quad 
        \tilde{c}_{j,l}=\begin{cases}
            c_{j+1} & \mbox{ if } l=1,\\
            c_j & \mbox{ if } l>1.
        \end{cases}
        \label{eq.ctilde}
    \end{align} 
    
    \item Similar to Step (v), we take all $c_{1}$ outputs from the V-cycle as inputs and use components from $A^{*}$ to convolve with them. We decompose $A^*$ as
		\begin{align}
			A^*(\xb,t)=\sum_{s=1}^{c_1} A_s^*(\xb,t),
		\end{align}
		where $A_s^*$ is used to convolve with the $s$-th output from level $1$ of the right branch of the V-cycle.
\end{enumerate}
 
After the decomposition, the control variables and operations are decomposed as:
	\begin{align}
		&A(\xb,t)=\sum_{j=1}^J \sum_{l=1}^{L_j}\sum_{k=1}^{c_j} \sum_{s=1}^{c_{j,l}}  A_{k,s}^{j,l}(\xb,t), \quad &&\tilde{A}(\xb,t)=\sum_{j=1}^J \sum_{l=1}^{L_j}\sum_{k=1}^{c_j} \sum_{s=1}^{\tilde{c}_{j,l}}  \tilde{A}_{k,s}^{j,l}(\xb,t)+ \sum_{s=1}^{c_1} A^*_s(\xb,t),  \label{eq.full.A}\\
		&b(t)=\sum_{j=1}^J \sum_{l=1}^{L_j}\sum_{k=1}^{c_j}   b_{k}^{j,l}(t), \quad &&\tilde{b}(t)=\sum_{j=1}^J \sum_{l=1}^{L_j}\sum_{k=1}^{c_j}  \tilde{b}_{k}^{j,l}(t)+  b^*(t),
		\label{eq.full.b}\\
		&S(u)=\sum_{j=1}^J  \sum_{l=1}^{L_j} \sum_{k=1}^{c_j} S_{k}^j (u), \quad && \tilde{S}(u)=\sum_{j=1}^{J-1}  \sum_{l=1}^{L_j} \sum_{k=1}^{c_j} \tilde{S}_{k}^j (u) +S^*(u).
		\label{eq.full.S}
	\end{align}
The original control problem  is transferred to minimize the loss (\ref{eq:loss} for $u$ being the solution of the following equation: 
	\begin{align}
		\begin{cases}
			\frac{\partial u}{\partial t}=A*u+\tilde{A}*u + b +\tilde{b} + S(u)+\tilde{S}(u), \ (\xb,t)\in \Omega\times [0,T],\\
			u(\xb,0)=H(f), \ \xb\in \Omega.
		\end{cases}
		\label{eq.control.full}
	\end{align}
From (\ref{eq.full.A})-(\ref{eq.full.b}), we see that the control variables $\theta_1=(W(x,t), b(t))$ are decomposed into a large sum and the items in these sums are the new control variables. The number of the control variables are large, but each of them is very small in number of unknowns. 
 
\begin{figure}
		\centering
		\includegraphics[width=\textwidth]{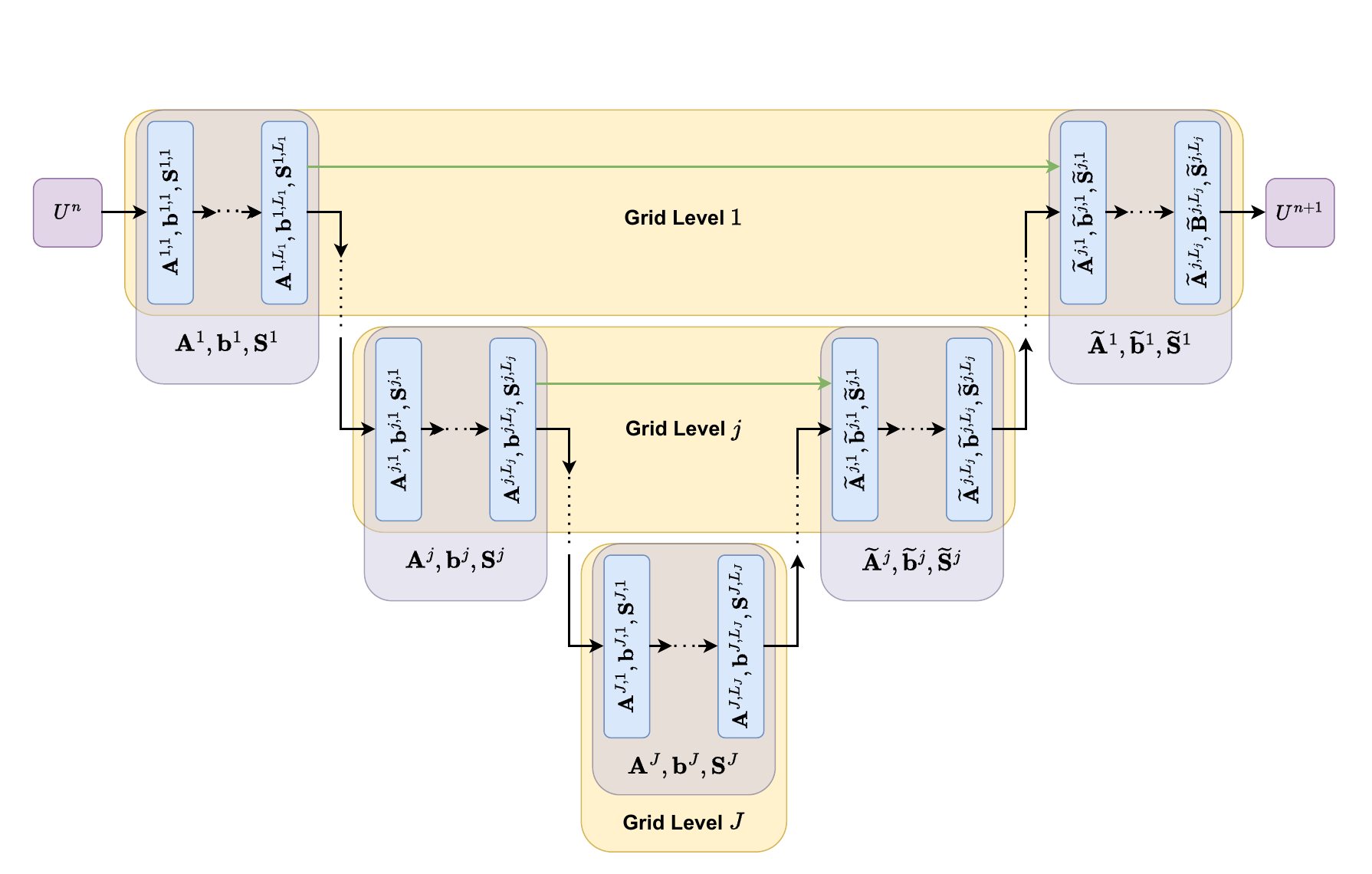}
		\caption{An illustration of Algorithm \ref{alg.V.full}.}
		\label{fig.alg}
	\end{figure}

To solve (\ref{eq.control.full}), we use the hybrid splitting method introduced in Section \ref{sec.hybrid}. Divide the time interval $[0,T]$ into $N$ subintervals with time step $\Delta t=T/N$. Denote the computed solution at time $t^n=n\Delta t$ by $U^n$. The resulting algorithm that updates $U^n$ to $U^{n+1}$ is summarized in Algorithm \ref{alg.V.full}. For simplicity, variable dependencies on $\xb$ are omitted. In Algorithm  \ref{alg.V.full}, we use $u^{j,l}, v^{j,l}$ to denote intermediate variables in the left and right branches, respectively. The superscript $j$ denotes the grid level at which the computation is conducted, and $l$ denotes the index of the sequential sub-step at grid level $j$. 
The architecture of Algorithm \ref{alg.V.full} is illustrated in Figure \ref{fig.alg}. In Figure \ref{fig.alg}, a relaxation step is used for each grid level to pass information from the left branch to the right branch, as indicated by the green arrows. The explanations of all indices for operators and variables of the left branch are summarized in Table \ref{tab.V.full}.

	\begin{algorithm}
		\caption{A hybrid splitting method to solve the control problem (\ref{eq.control.full})}
		\label{alg.V.full}
		\begin{algorithmic}
			\STATE {\bf Data:} The solution $U^n$ at time $t^n$.
			\STATE {\bf Result:} The computed solution $U^{n+1}$ at time step $t^{n+1}$.
			
			{\bf Set} $c_{0} = 1, L_0=1, v^{1,0}=v^{1,0}_1=U^n$. \\
			\FOR {$j = 1,  \cdots, J$}
			\STATE 
			If $j>1$, set $v^{j,0}=\cD(v^{j-1,L_{j-1}})$ and $v_k^{j,0}=\cD(v_k^{j-1,L_{j-1}})$ for $k=1,...,c_{j-1}$.
			\FOR {$l=1,...,L_{j}$}
			\FOR {$k=1,...,c_{j}$}
			\STATE Compute $v_k^{j,l}$ on $\cV^j$ by solving
			\begin{align}
				\frac {v_k^{j,l} - v^{j,l-1}} {2^{j-1}c_j \Delta t } -   
				\sum_{s=1}^{c_{j,l}} 
				A_{k,s}^{j,l}(t^n)  * v_s^{j,l-1}   - b_{k}^{j,l}(t^n)   - S_{k}^{j,l}(v_k^{j,l} ) \ni 0,
				\label{eq.full.v}
			\end{align}
			where $c_{j,l}$ is defined in (\ref{eq.c}).
			\ENDFOR 
			
			Compute $v^{j+1,l}$ as
			\begin{equation*}
				v^{j,l} = \frac 1 {c_{j}} \sum_{k=1}^{c_{j}} v_k ^ {j,l} . 
			\end{equation*}
			\ENDFOR 
			
			\ENDFOR
			
			{\bf Set} $u^{J,L_J}= v^{J,L_J}$ and $ u_k^{J,L_J} = v_k^{J,L_J}$ for $k =1, 2, \cdots c_J$.
			\FOR {$j = J-1,  \cdots, 1$}
			\STATE Set $u^{j,0}=\cU(u^{j+1,L_{j+1}})$ and for $k=1,...,c_{j+1}$, compute
			$$u_k^{j,0}=\frac{1}{2}u_k^{j+1,L_{j+1}}+ \frac{1}{2}\cU(u^{j+1,L_{j+1}})$$
			\FOR {$l=1,...,L_j$}
			\FOR {$k=1,2,\cdots c_j$}
			\STATE Compute $u_k^{j,l}$ on $\cV^j$ by solving 
			\begin{align}
				&\frac {u_k^{j,l} - u^{j,l-1}} {2^{j-1} \tilde{c}_{j} \Delta t } -   
				\sum_{s=1}^{\tilde{c}_{j,l}} 
				\tilde A_{k,s}^{j}(t^n)  * u_s^{j,l-1}   - \tilde b_{k}^{j}(t^n)  - \tilde S_{k}^j(u_k^{j} ) \ni 0,
				\label{eq.full.u}
			\end{align}					
			where $\tilde{c}_{j,l}$ is defined in (\ref{eq.ctilde}).
			\ENDFOR
			
			Compute $u^{j,l}$ as
			\[
			u^{j,l} = \frac 1 {c_{j}} \sum_{k=1}^{c_j} u_k ^ {j,l}.
			\]
			
			\ENDFOR

			\ENDFOR
			
			Compute $U^{n+1}$ by solving
			\begin{align}
				\frac {U^{n+1} - u^{1,L_1}} {\Delta t } -   
				\sum_{s=1}^{c_{1}} 
				A_{s}^{*}(t^n)  * u_s^{1,L_1}   - b^*(t^n)   -  S^*(U^{n+1} )\ni 0.
				\label{eq.full.final}
			\end{align}
		\end{algorithmic}
	\end{algorithm}
	
\begin{table}[t!]
	\centering
	\begin{tabular}{c|c|c|c|c}
		\hline
		\makecell{For $A^{j}_{k,s},b^{j}_{k},S^{j}_k$,\\ $A^{j,l}_{k,s},b^{j,l}_{k},S^{j,l}_k$} & $j$ & $l$ & $k$ & $s$\\
		\hline
		\makecell{ Index meaning: \\index of} & grid levels & \makecell{sequential \\splittings} & \makecell{parallel \\splittings} & \makecell{output from \\ the previous substep}\\
		\hline
		\makecell{For $u^{j}_k,v^{j}_k$,\\ $u^{j,l}_k,v^{j,l}_k$} &$j$ & $l$ & $k$ &-\\
		\hline
		\makecell{ Index meaning: \\index of} & grid levels & \makecell{sequential \\splittings} & \makecell{parallel \\splittings} & -\\
		\hline
	\end{tabular}
	\caption{Explanation of indices for kernels and variables in the left branch of \ref{alg.V.full}.}
	\label{tab.V.full}
\end{table}

	Denote $\theta_2=\{\theta_2^n\}_{n=1}^N$ with
\begin{align*}
	\theta_2^n =\big(&\{ A^{j,l}_{k,s}(\xb, t^n)\}_{j,l,k,s},\{ \tilde A^{j,l}_{k,s}(\xb, t^n)\}_{j,l,k,s}, \{A_s^*(\xb, t^n)\}_s, \\
	&\{b_k^{j,l}( t^n)\}_{j,l,k}, \{ \widetilde{b}_k^{j,l}( t^n)\}_{j,l,k}, \widetilde{b}^*( t^n)\big).  
	\label{eq.theta2}
\end{align*}
We also denote $\cN_2$ as the mapping:
$$
\cN_2:f\rightarrow H(f)\rightarrow U^1 \rightarrow \cdots \rightarrow U^N, 
$$
which maps $f$ to $U^N$ by applying Algorithm \ref{alg.V.full} $N$ times. Parameters in $\theta_2$ are learned by solving
\begin{equation}
	\min_{\theta_2} \sum_{i=1}^I \mathcal{L}(\mathcal{N}_2(f_i,\theta_2) , g_i ).
	\label{eq.min.basic}
\end{equation}
In (\ref{eq.min.basic}), $\theta_2$ is a space decomposition representation for a discretization of $\theta_1$. The operation procedure $\cN_2$ is a numerical scheme solving (\ref{eq.control0}). We can see that problem (\ref{eq.min.basic}) is a discretization of the optimization problem (\ref{eq:loss}) with some proper decomposition of the control variables.

	\section{Algorithm details}
	\label{sec.implementation}
	
	In Algorithm \ref{alg.V.full}, one needs to solve (\ref{eq.full.v}), (\ref{eq.full.u}) and (\ref{eq.full.final}), which includes components of $S,\tilde{S}$. We discuss the choices of $S,\tilde{S}$ and present how to solve (\ref{eq.full.v}), (\ref{eq.full.u}) and (\ref{eq.full.final}) in the following subsections.
	\subsection{On the choices of $S,\tilde{S}$}
	\label{sec.select}

	According to (\ref{eq.S.sum}), $S+\tilde{S}$ consists of two terms: (i) The first term is $-\ln \frac{u}{1-u}$, which will be used in $S^*$. This term enforces u to be between 0 and 1 and provides the prediction results. (ii) The second term $-\partial \cI_{\Sigma}(u)$ will be used at every sub-step except for $S^*$. We will show that this part corresponds to the ReLU activation function in a network. Specifically, we set 
	\begin{align}
		S^{j,l}_k(u)= \tilde{S}^{j,l}_k(u)= \partial \cI_{\Sigma}(u) , S^*(u)=-\ln  \frac u {1-u}.
		\label{eq.S.choice}
	\end{align}

	\subsection{On the solution to (\ref{eq.full.v}), (\ref{eq.full.u}) and (\ref{eq.full.final})}
	\label{sec.solver}

	Observe that (\ref{eq.full.v}) and (\ref{eq.full.u}) are in the form of
	\begin{align}
		\frac{u-u^*}{\gamma \Delta t}-\sum_{s=1}^c \hat{A}_s*u^*_s -\hat{b} +\partial \cI_{\Sigma} (u)\ni 0,
		\label{eq.basic.form}
	\end{align}
	where $\gamma$ is some constant, $c$ is some integer, $u^*=\frac{1}{c} \sum_{s=1}^c u^*_s$ for some functions $u^*_s$'s, $\hat{A}_s$'s are some convolution kernels, $\hat{b}$ is some bias function. The solution to (\ref{eq.basic.form}) can be computed using the following two-sub-step splitting method:
	\begin{align}
		\begin{cases}
			\bar{u}=u^*+\gamma\Delta t \left(\sum_{s=1}^c \hat{A}_s*u^*_s +\hat{b}\right),\\
			\frac{u-\bar{u}}{\gamma \Delta t}+\partial \cI_{\Sigma} (u)\ni 0.
		\end{cases}
		\label{eq.basic.form.sub}
	\end{align}	
	
	In (\ref{eq.basic.form.sub}), there is no difficulty in solving for $\bar{u}$ in the first sub-step as it is an explicit step. For u in the second sub-step, it is, in fact, a projection. Its closed-form solution is given as
	\begin{align}
		u=\max\{\bar{u},0\}=\ReLU(\bar{u}),
	\end{align}
where $\ReLU(u)=\max\{\bar{u},0\}$ is the rectified linear unit.
	
	Problem (\ref{eq.full.final}) can be written as 
	\begin{align}
		\frac{u-u^*}{\gamma \Delta t}=\sum_{s=1}^c \hat{A}_s*u^*_s +\hat{b}- \ln \frac{u}{1-u}.
		\label{eq.basic.form.final}
	\end{align}
	Following the steps for solving (\ref{eq.full.v}) and (\ref{eq.full.u}) above, we solve (\ref{eq.basic.form.final}) as
	\begin{align}
		\begin{cases}
			\bar{u}=u^*+\gamma\Delta t \left(\sum_{s=1}^c \hat{A}_s*u^*_s +\hat{b}\right),\\
			\frac{u-\bar{u}}{ \Delta t}=-\ln \frac{u}{1-u}.
		\end{cases}
		\label{eq.basic.form.final.sub}
	\end{align}	
	The first sub-step is an explicit step. We solve the second sub-step approximately by a fixed point iteration.
	
	Initialize $p^0=\bar{u}$. Given $p^k$, we update $p^{k+1}$ by solving
	\begin{align}
		\frac{p^k-\bar{u}}{\Delta t}=-\ln \frac{p^{k+1}}{1-p^{k+1}},
	\end{align}
	for which we have the closed-form solution
	\begin{align}
		p^{k+1}=\Sig\left(-\frac{p^k-\bar{u}}{\Delta t} \right),
		\label{eq.p.update}
	\end{align}
	where $\Sig(x)=\frac{1}{1+e^{-x}}$ is the sigmoid function. By repeating (\ref{eq.p.update}) so that $p^{k+1}$ converges to some function $p^*$, we set $u=p^*$.
	In particular, since $p^0=\bar{u}$, the updating formula (\ref{eq.p.update}) always gives $p^1=0.5$. If we only consider a two-step fixed point iteration, we get 
	\begin{align}
		u=\Sig\left(-\frac{0.5-\bar{u}}{\Delta t} \right)=\Sig\left(\frac{\bar{u}-0.5}{\Delta t} \right).
	\end{align} 
	
	\subsection{Initial condition}
	Problem (\ref{eq.control}) requires an initial condition. A simple choice is to set it as some convolution of $f$:
	\begin{align}
		u(\xb,0)=H(f)=\Sig\left(\sum_{k=1}^3 A^0_k\ast f^k\right)
	\end{align}
for $f=(f^1,f^2,f^3)$. $f^1$, $f^2$, and $f^3$ denote the RGB channels of an image respectively.

 \subsection{Discretization }
 To discretize a continuous function $u$ at grid level $j$, we compute the scaled inner product
 $$
 a_{\balpha}^j= \frac{1}{(2^{j-1}h)^2}\int_{\Omega} u\phi_{\balpha}^jdxdy
 $$
 for each basis function $\phi_{\balpha}^j$ (defined in (\ref{eq.basis})) of the space $\cV^j$. Note that each $\phi_{\alpha}^j$ is an indicator function of a $(2^{j-1}h\times 2^{j-1}h)$ patch indexed by $\balpha$. The inner product $a_{\balpha}^j$ gives the pixel value of $u$ at the $\balpha$-th patch. We take the original image resolution as grid level 1 (the finest grid). Other grid levels and the basis functions can be defined according to the discussion in Section \ref{sec.multigrid}.

	\section{Algorithm \ref{alg.V.full} recovers UNet}  
	\label{sec.networks}
	In this section, we show that by properly setting the number of grid levels $J$, parallel splittings $c_j$'s, and sequatial splittings $L_j$'s, Algorithm \ref{alg.V.full} exactly recovers UNet.
	\subsection{Algorithm \ref{alg.V.full} building blocks recover UNet layers}
	\label{sec.buildingblock}
We first show that a building block of Algorithm \ref{alg.V.full} is equivalent to a layer of UNet. Each layer of UNet is a convolution layer activated by ReLU. Given outputs from the previous layer $\{v_s^*\}_{s=1}^c$, a UNet layer outputs $v$ by the following operations:
	\begin{align}
		\begin{cases}
			\bar{v}=\sum_{s=1}^c W_{s}*v_s^*+b,\\
			v=\ReLU(\bar{v}),
		\end{cases}
		\label{eq.network.block}
	\end{align}
	where $W_{s}$'s are convolutional kernels and $b$ is the bias. In Algorithm \ref{alg.V.full}, the building block is (\ref{eq.basic.form}) and (\ref{eq.basic.form.final}), which is solved by (\ref{eq.basic.form.sub}) and (\ref{eq.basic.form.final.sub}).  In fact, (\ref{eq.network.block}) (or problem (\ref{eq.basic.form.final.sub})) and (\ref{eq.basic.form.sub}) have the same form.

	Specifically, in the first equation of (\ref{eq.basic.form.sub}), substitute the expression of $u^*$, and we have
	\begin{align}
		\bar{u}=&\frac{1}{c}\sum_{s=1}^c u^*_s + \gamma\Delta t\left(\sum_{s=1}^c \hat{A}_s*u_s^* +\hat{b}\right)=\sum_{s=1}^c \left(\frac{1}{c} \mathds{1}+\gamma\Delta t \hat{A}_s\right)*u_s^* + \gamma\Delta t\hat{b},
	\end{align}
	where $\mathds{1}$ denotes the identity kernel satisfying $\mathds{1}*g=g$ for any function $g$. In (\ref{eq.network.block}), set 
	\begin{align}
		W_{s}=\frac{1}{c} \mathds{1}+\gamma\Delta t \hat{A}_s, \quad b=\hat{b}.
	\end{align}
	We have $\bar{v}=\bar{u}$, and $v=u$. Essentially, Algorithm \ref{alg.V.full} and UNet have the same building block. 
	
	\subsection{Algorithm \ref{alg.V.full} structure recovers UNet architecture}
        UNet architecture consists of four components: encoder, decoder, bottleneck and skip-connections, each of which has a corresponding component in the structure of Algorithm \ref{alg.V.full}:
	\begin{enumerate}[label=(\roman*)]
		\item {\bf Encoder}: Encoder in UNet corresponds to the left branch of the V-cycle in Algorithm \ref{alg.V.full}. The number of data resolution levels corresponds to the number of grid levels $J$. At each data resolution, the number of layers and the width of each layer correspond to the number of sequential splittings $L_j$ and parallel splittings $c_j$ at the corresponding grid level.

        \item {\bf Decoder}: Decoder in UNet corresponds to the right branch of the V-cycle in Algorithm \ref{alg.V.full}. The number of data resolution levels corresponds to the number of grid levels $J$. At each data resolution, the number of layers and the width of each layer correspond to the number of sequential splittings $L_j$ and parallel splittings $c_j$ at the corresponding grid level.

        \item {\bf Bottleneck}: Bottleneck in UNet corresponds to the computations at the coarsest grid level (grid level $J$) in Algorithm \ref{alg.V.full}. The number of layers and layer width in bottleneck correspond to the number of sequential splittings $L_J$ and parallel splitting $c_J$ at grid level $J$.

		\item {\bf Skip-layer connection}: Skip-layer connections in UNet correspond to the relaxation steps in Algorithm \ref{alg.V.full}.

	\end{enumerate}

        UNet has 5 data resolution levels. For each resolution level, there are two layers in the encoder, decoder and bottleneck. From the finest resolution to the coarsest resolution, the layers has width $64,128,256,512,1024$. Thus,  set $J=5$,  $L_1=L_2=L_3=L_4=L_5=2$, $[c_1,c_2,c_3,c_4,c_5]=[64,128,256,512,1024]$, downsampling operator $\cD$ as the max-pooling operator, and upsampling operator $\cU$ as the transpose convolution, Algorithm \ref{alg.V.full} exactly recovers UNet. As a consequence, one can explain UNet as a one-step operator-splitting algorithm solving a control problem (\ref{eq.control0}).

	\section{Conclusion}
	\label{sec.conclusion}
	In this paper, we consider the control problem (\ref{eq.control0}) and propose an operator-splitting method to solve it. The ingredients of our algorithm include the multigrid method and the hybrid operator splitting method. We show that the resulting algorithm has the same building block and architecture as UNet. Our result demonstrates that UNet is a one-step operator-splitting algorithm that solves some control problems; thus, it gives a mathematical explanation of the UNet architecture from an algorithmic perspective.

    \bibliographystyle{abbrv}
    \bibliography{ref}

\end{document}